\documentclass[letterpaper, 10 pt, conference]{ieeeconf}  %
\pdfoutput=1

\IEEEoverridecommandlockouts       

\usepackage[backend=biber,
            url=false,
            isbn=false,
            doi=false,
            backref=false,
            style=ieee,
            natbib=true,%
            mincitenames=1,
            maxcitenames=1,
            citestyle=numeric-comp,
            sorting=nyt,%
            block=none]{biblatex}

\renewcommand{\bibfont}{\small}

\usepackage{algorithm}
\usepackage[noend]{algpseudocode}

\usepackage{graphics}
\usepackage{bbm}
\usepackage[pdftex]{graphicx}
\usepackage{wrapfig}
\DeclareGraphicsExtensions{.pdf,.png,.jpg}
\usepackage{epsfig}
\usepackage[font={small}]{caption}
\usepackage{subcaption}
\usepackage[rightcaption]{sidecap}
\usepackage{pbox}

\usepackage{bigstrut}
\usepackage{mathtools}
\usepackage{amsmath, amssymb, amscd}
\usepackage{ wasysym } %
\usepackage{amsfonts}
\usepackage{mathptmx} %
\usepackage{gensymb} 
\usepackage{nicefrac}       %
\numberwithin{equation}{section} 

\DeclareMathAlphabet{\mathcal}{OMS}{lmsy}{m}{n}
\DeclareSymbolFont{largesymbols}{OMX}{cmex}{m}{n}
\usepackage{textcomp} %

\usepackage{algorithm} %
\usepackage{algorithmicx, algpseudocode} %

\usepackage{array} %
\usepackage{tabularx}
\usepackage{multirow}
\usepackage{booktabs}
\usepackage{tabulary}

\usepackage[T1]{fontenc} 
\usepackage[utf8]{inputenc}
\usepackage[english]{babel} %
\usepackage{units}
\usepackage{bm}
\usepackage{times} %
\usepackage{xspace}
\usepackage{flushend}%
\usepackage{balance} 
\usepackage{csquotes}
\usepackage{makeidx}
\usepackage{blindtext}

\let\labelindent\relax
\usepackage{enumitem}

\usepackage{ragged2e}
\usepackage{soul} %
\usepackage{subfiles} %

\usepackage[protrusion=true,expansion=true]{microtype}
\usepackage[yyyymmdd]{datetime}

\date{\protect\formatdate{1}{1}{2001}}

\usepackage{url}
\makeatletter
\g@addto@macro{\UrlBreaks}{\UrlOrds}
\makeatother
\usepackage{color}
\usepackage[usenames,dvipsnames,table,xcdraw]{xcolor}
\usepackage{pgfplots} %
\pgfplotsset{compat=newest}

\usepackage{marginnote}
\usepackage{soul} %

\usepackage[colorinlistoftodos]{todonotes}
\usepackage{multicol}
\newcommand{\tocite}[1]{%
\textcolor{red}{[cite:\ifthenelse{\equal{#1}{}}{}{#1}?]}
}

\newcommand{\ignore}[1]{}

\renewcommand{\baselinestretch}{1.0}



\DeclareMathOperator*{\argmin}{arg\,min}

\newcommand{\ba}{\mathbf{a}}

\title{\LARGE \bf
Learning Dense Visual Correspondences \\ in Simulation to Smooth and Fold Real Fabrics
}

\author{Aditya Ganapathi$^{1}$, Priya Sundaresan$^{1}$, Brijen Thananjeyan$^{1}$, Ashwin Balakrishna$^{1}$, \\ Daniel Seita$^{1}$, Jennifer Grannen$^{1}$, Minho Hwang$^{1}$, Ryan Hoque$^{1}$, \\ Joseph E. Gonzalez$^{1}$, Nawid Jamali$^2$, Katsu Yamane$^2$, Soshi Iba$^2$, Ken Goldberg$^{1}$
\thanks{$^{1}$AUTOLab at the University of California, Berkeley, USA}
\thanks{$^{2}$Honda Research Institute, USA}
\thanks{Correspondence to Aditya Ganapathi: \texttt{avganapathi@berkeley.edu}}
}

\begin{document}

\maketitle

\begin{abstract}
Robotic fabric manipulation is challenging due to the infinite dimensional configuration space, self-occlusion, and complex dynamics of fabrics. There has been significant prior work on learning policies for specific deformable manipulation tasks, but comparatively less focus on algorithms which can efficiently learn many different tasks. In this paper, we learn visual correspondences for deformable fabrics across different configurations in simulation and show that this representation can be used to design policies for a variety of tasks. Given a single demonstration of a new task from an initial fabric configuration, the learned correspondences can be used to compute geometrically equivalent actions in a new fabric configuration. This makes it possible to robustly imitate a broad set of multi-step fabric smoothing and folding tasks on multiple physical robotic systems. The resulting policies achieve $80.3\%$ average task success rate across 10 fabric manipulation tasks on two different robotic systems, the da Vinci surgical robot and the ABB YuMi. Results also suggest robustness to fabrics of various colors, sizes, and shapes. See \url{https://tinyurl.com/fabric-descriptors} for supplementary material and videos.

\end{abstract}

\section{Introduction}
\label{sec:introduction}
Robot fabric manipulation has applications in folding laundry~\cite{willimon_unfolding_laundry_2011, kita_2009_icra,unfolding_rf_2014, laundry2012}, bed making~\cite{bed-making}, surgery~\cite{thananjeyan2017multilateral, SAVED,rosen_icra_tissues_2019, rope-untangle}, and manufacturing~\cite{10.1115/1.3185859, Torgerson1987VisionGR}. However, while robots are able to learn general purpose policies to manipulate a variety of rigid objects with increasing reliability~\cite{dexnet4, dense-obj-nets, pinto2015supersizing, qt-opt, hand-eye}, learning such policies for manipulating deformable objects remains an open problem due to difficulties in sensing and control. While there is significant prior work on geometric~\cite{schulman_isrr_2013, willimon_unfolding_laundry_2011, balaguer2011combining, maitin2010cloth} and learning based approaches~\cite{lerrel, seita_ryan, bed-making} for fabric manipulation, these approaches often involve designing or learning task-specific manipulation policies, making it difficult to efficiently reuse information for different tasks.

In this work, we leverage recent advances in dense keypoint learning~\cite{dense-obj-nets} to learn visual point-pair correspondences across fabric in different configurations. Then, given a single offline demonstration of a fabric manipulation task from a given configuration, we utilize the learned correspondences to compute geometrically equivalent actions to complete the task on a similar fabric in a different configuration. For example, a human might provide a sequence of actions that would fold a T-shirt when it is placed neck up in a smoothed configuration. However, when a robot is operating at test time, it will likely encounter a different T-shirt whose color, size and pose may differ from the T-shirt used for the demonstration. We find that learning visual correspondences that are invariant across these fabric attributes provides a powerful representation for defining controllers that can generalize to the above variations.

\begin{figure}[t]
\center
\includegraphics[width=\columnwidth]{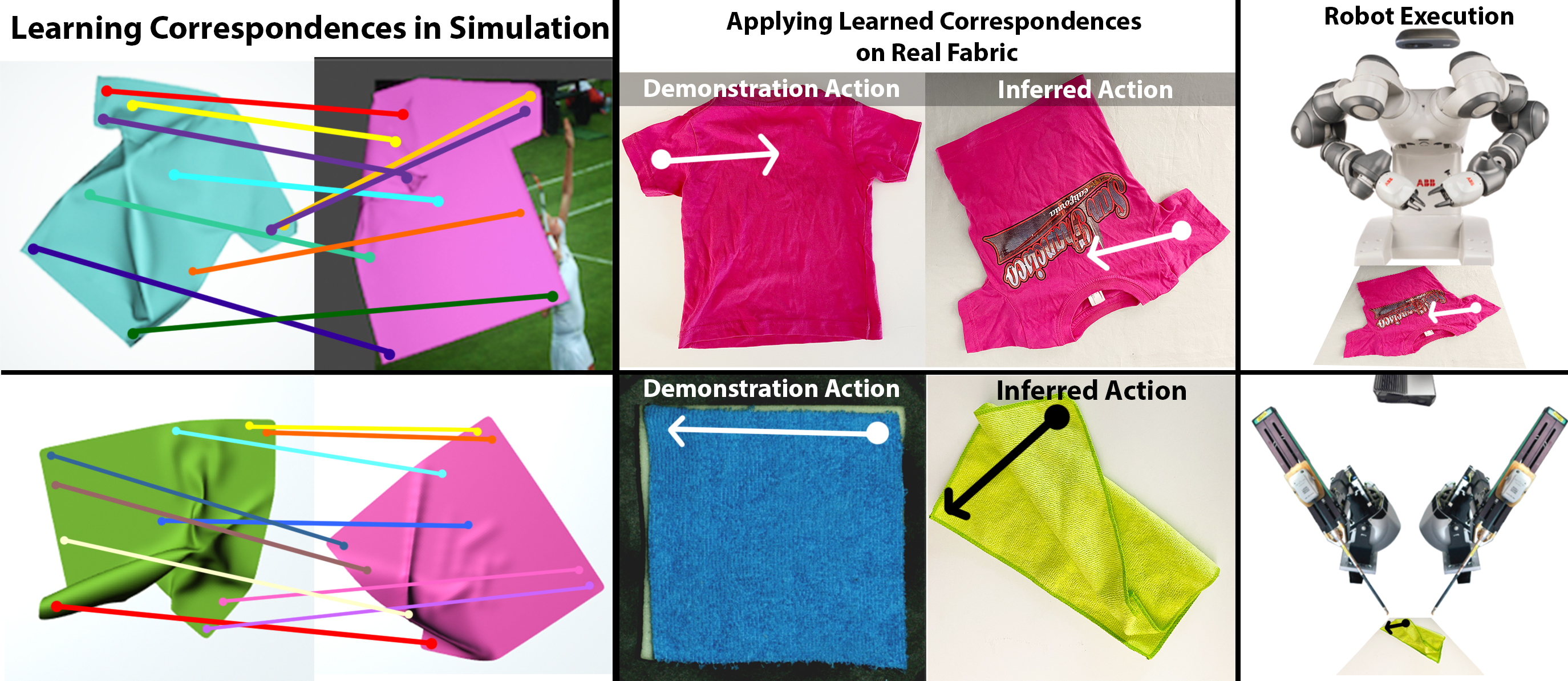}
\caption{
We use learned visual correspondences across different fabric configurations to perform a variety of fabric manipulation tasks on  the ABB YuMi (top) and the da Vinci Research Kit (bottom). Given a single demonstration of smoothing or folding, the robot uses the learned correspondences to compute geometrically equivalent actions for fabric of different color and in different initial configurations. This enables robust one-shot imitation learning of tasks that involve smoothing then folding.
}
\label{fig:teaser}
\end{figure}

We extend work by~\citet{priya-rope}, which leverages dense object descriptors~\cite{dense-obj-nets} to learn visual correspondences for rope using synthetic depth data in simulation. These correspondences are then used to learn new rope manipulation tasks such as rearrangement or knot tying given a single task demonstration. We find that similar visual correspondence learning methods are also effective for learning correspondences between different fabric configurations using task-agnostic RGB data collected entirely in simulation and can be used to perform fabric manipulation tasks. Precisely, given a user demonstration of the task from a given initial fabric configuration, we leverage the learned visual correspondences to perform the same task from different initial configurations by computing geometrically equivalent actions using the correspondences. This approach has a number of appealing properties. First, visual correspondences can be learned purely in simulation without task-specific data and widely applied to a variety of real fabric manipulation tasks with no further training. Second, training in simulation enables sufficient data variety through domain randomization, making it possible to learn correspondences that generalize to fabrics with different colors, shapes, and configurations. Third, since perception and control are decoupled, the same perception module can be used on different robots with no additional training.

We contribute (1) a framework for learning dense visual correspondences of fabric in simulation using dense object descriptors from~\cite{dense-obj-nets,priya-rope} and applying them to manipulation tasks on real fabrics with unseen colors, scales, and textures,
(2) a data generation pipeline for collecting images of fabrics and clothing in Blender~\cite{blender} and a testbed to experiment with different manipulation policies on these fabrics in simulation
and (3) physical experiments on both the da Vinci Research Kit (dVRK)~\cite{dvrk2014} and the ABB YuMi suggesting that the learned descriptors transfer effectively on two different robotic systems. We experimentally validate the method on 10 different tasks involving 5 T-shirts and 5 square fabrics of varying dimensions and colors and achieve an average task success rate of $80.3\%$.
\section{Related Work}
\label{sec:related-work}
Fabric manipulation is an active area of robotics research~\cite{grasp_centered_survey_2019,manip_deformable_survey_2018,robot-learning-manip-2019, rishabh_2019}. Over the past decade, the research has primarily been focused on three different categories: perception-based manipulation, learning-based algorithms in the real world, and learning-based algorithms in simulation which are then transferred to real robots.

\textbf{Traditional Vision-Based Algorithms for Fabric Manipulation:}  Much of the prior work on perception-based deformable object manipulation relies on traditional image processing and vision techniques to estimate the state of the fabric. This state estimation is then used to define geometric controllers which bring the fabric into some desired configuration. However, due to limitations in these traditional vision algorithms, most prior work makes specific assumptions on the fabric's initial configurations or requires more complex robotic manipulators to bring the fabric into a desired starting configuration. For example, Miller~\emph{et~al}.~\cite{laundry2012} demonstrate a robust folding pipeline for clothing by fitting a polygonal contour to the fabric and designing a geometric controller on top of it, but assume that the initial state of the fabric is flat. Sun~\emph{et~al}.~\cite{heuristic_wrinkles_2014, cloth_icra_2015} perform effective fabric smoothing by estimating the wrinkles in the fabric, but condition on a near-flat starting fabric. Other work relies on ``vertically smoothing'' fabrics using gravity~\cite{osawa_2007,kita_2009_iros,kita_2009_icra,unfolding_rf_2014, maitin2010cloth} to standardize the initial configuration and to expose fabric corners before attempting the task, which is difficult for large fabrics or single-armed robots.

\textbf{Learning-Based Algorithms in the Real World:}
More recent approaches have transitioned to end to end learning of fabric manipulation directly on a real system, but these approaches have struggled to generalize to a variety of fabrics and tasks due to the high volume of training data required. For example, \citet{visual_foresight_2018} use model-based reinforcement learning to learn fabric manipulation policies which generalize to many tasks, but require several days of continuous data collection on a real physical system and perform relatively low precision tasks. Jia~\emph{et~al}.~\cite{jia_visual_feedback_2018,jia_cloth_manip_2019} show impressive collaborative human-robot cloth folding under the assumption that fabric has already been grasped and is in a particular starting configuration, and~\citet{schulman_isrr_2013} demonstrate deformable object manipulation while requiring task-specific kinesthetic demonstrations. In follow-up work,~\citet{fabric_folding_real} consider many of the same tasks as in this paper and demonstrate that policies can be learned to fold fabric using reinforcement learning with only one hour of experience on a real robot. In contrast, we learn entirely in simulation and decouple perception from control, making it easier to generalize to different fabric colors and shapes and flexibly deploy the learned policies on different robots.

\textbf{Sim-to-Real Learning-Based Algorithms:} Due to the recent success of sim-to-real transfer~\cite{cad2rl,domain_randomization}, many recent papers leverage simulation to collect large amounts of training data which is used to learn fabric manipulation policies. Seita~\emph{et~al}.~\cite{bed-making,seita_ryan} and ~\citet{lerrel} followed-up on the smoothing task from \cite{heuristic_wrinkles_2014} by generalizing to a wider range of initial fabric states using imitation learning (DAgger~\cite{ross2011reduction}), and reinforcement learning (Soft Actor-Critic~\cite{sac}) respectively, but still tailor policies specifically for smoothing. Similarly,~\citet{sim2real_deform_2018} learn fabric folding policies by using deep reinforcement learning augmented with task-specific demonstrations. These works use simulation to optimize fabric manipulation policies for specific tasks. In follow-up and concurrent work,~\citet{fabric_vsf_2020} and~\citet{fabric_latents_2020} use simulation to train fabric manipulation policies using model-based reinforcement learning for multiple tasks. In contrast, we leverage simulation to learn visual representations of fabric to capture its geometric structure without task-specific data or a model of the environment and then use this representation to design intuitive controllers for several tasks from different starting configurations. 

\textbf{Dense Object Descriptors:} We learn visual representations for fabric by using dense object descriptors~\cite{dense-obj-nets, visual-descriptors}, which were shown to enable task oriented manipulation of various rigid and slightly deformable objects~\cite{dense-obj-nets}. This approach uses a deep neural network to learn a representation which encourages corresponding pixels in images of an object in different configurations to have similar representations in embedding space. Such descriptors can be used to design geometrically structured manipulation policies for grasping~\cite{dense-obj-nets}, assembly~\cite{form2fit}, or for learning from demonstrations~\cite{descriptors_2020}. ~\citet{priya-rope} extend this idea to manipulation of ropes, and demonstrate that deformation-invariant dense object descriptors can be learned for rope using synthetic depth data in simulation and then transferred to a real physical system.~\citet{priya-rope} then use the learned descriptors to imitate offline demonstrations of various rope manipulation tasks. In this work, we apply the techniques from \cite{priya-rope} to learn descriptors which capture geometric correspondence across different fabric configurations from synthetic RGB images and use them for 2D fabric manipulation.
\section{Problem Definition}
\label{sec:problem_definition}
\subsection{Assumptions}
\label{subsec:assumptions}
We assume a deformable object is place on a planar workspace in an initial configuration $\xi_1$ observed by an overhead camera corresponding RGB image $I_1 := I_1(\xi_1) \in \mathbb{R}^{W \times H \times 3}$. As in prior work~\cite{seita_ryan, lerrel}, we assume that fabric manipulation tasks can be completed by a sequence of actions where each includes grasping at a \emph{pick point}, pulling to a \emph{place point} without changing the orientation of the end-effector, and releasing at the place point.
We additionally assume access to a single demonstration of each task in the form of a sequence of pick and place actions from some arbitrary initial fabric configuration $\xi_1$. These demonstrations can be collected offline, such as through a GUI where a user clicks on an image of fabric to indicate pick and place point pixels. However, the fabric used to create the instruction does not have to be of the same color, the same size or in the same initial configuration as the fabric the robot sees at test time. The only requirement is that the fabric be of a similar geometry. For example, T-shirts can be compared to other instances of T-shirts, but not to pants or long-sleeved shirts.
\subsection{Task Definition}
\label{subsec:task_definition}
Define the action at step $j$ as
\begin{equation}
\ba_{j} = ((x_g, y_g)_j, (x_p, y_p)_j)
\end{equation}
where $(x_g, y_g)_j$ and $(x_p, y_p)_j$ are the pixel coordinates of a grasp point on the fabric and place point respectively in image $I_j$ at time $j$. The robot grasps the world coordinate associated with the grasp point and then moves to the world coordinate associated with the place point without changing the end effector orientation. This causes the fabric located at $(x_g, y_g)_j$ in the image to be placed on top of the world coordinate associated with $(x_p, y_p)_j$ with the same surface normals as before. In future work, we will investigate how to execute more complex actions that result in reversed surface normals, which requires a rotation motion during the action.
We are given a sequence of actions $\left( \ba_j \right)_{j=1}^{n}$ executed on a fabric starting in configuration $\xi_1$ and corresponding observations $\left( I_j \right)_{j=1}^{n}$. Then at test-time, a similar object is dropped onto the surface in a previously unseen configuration and the goal is to generate a corresponding sequence of actions for a fabric in some previously unseen configuration. Specifically, the robot generates a new sequence of actions:

\begin{equation}
\Big(\ba_j'\Big)_{j=1}^{n} =  \Big( d_{I_j \rightarrow I_j'}(  x_g, y_g )_j, \;\; d_{I_j \rightarrow I_j'}(  x_p, y_p )_j \Big)_{j=1}^n
\end{equation}

for $j \in \{1, \ldots, n\}$ where $d_{I_j \rightarrow I_j'} : \mathbb{R}^2 \rightarrow \mathbb{R}^2$ is a function which estimates the corresponding point $(x', y')_j$ in $I_j'$ given a point $(x, y)_j$ in $I_j$. This function is difficult to compute directly from images in general, and even more so difficult to compute for images of highly deformable objects due to their infinite degrees of freedom and tendency to self-occlude. Thus, we leverage dense object descriptors~\cite{dense-obj-nets} to approximate $d_{I_j \rightarrow I_j'}$ for any $I_j$ and $I_j'$, as described in Sections~\ref{sec:descriptors} and~\ref{sec:policy_design}.

\section{Simulator}
\label{sec:simulator}

We use Blender 2.8, an open-source simulation and rendering engine~\cite{blender} released in mid-2019, to both create large synthetic RGB training datasets and model the fabric dynamics for simulated experiments using its in-built fabric solver  %
based on~\cite{provot1995deformation, provot1997collision}.
We simulate T-shirts and square fabrics, each of which we model as a polygonal mesh made up of 729 vertices, a square number we experimentally tuned to trade-off between fine-grained deformations and reasonable simulation speed. These meshes can be easily constructed in Blender by starting with a planar grid of vertices and then removing vertices and edges to create desired shapes. See Figure~\ref{fig:blender} for an illustration.
Each vertex on the mesh has a global coordinate which we can query directly through Blender's API, allowing for easily available ground truth information about various locations on the mesh and their pixel counterparts. See Figure~\ref{fig:blender} in the supplement for examples of these meshes. We can also simulate finer-grained manipulation of the mesh including grasps, pulls, and folds. To implement the action space defined in Section~\ref{sec:problem_definition}, we first deproject the pixel corresponding to the pick point and map it to the vertex whose global coordinates are closest in $\mathbb{R}^3$ to the pixel's deprojected coordinates. We then directly manipulate this vertex by pinning it and translating it over a sequence of 30 frames. This generalized form of manipulation allows us to easily execute experiments in simulation. See the supplement for further details on how we pin vertices, the overall technique and experiments in simulation.

\section{Dense Shape Descriptor Training}
\label{sec:descriptors}

\subsection{Dense Object Descriptor Training Procedure}
\label{subsec:preliminaries}
We consider an environment with a deformable fabric on a flat tabletop and learn policies that perform smoothing and folding tasks. The policies we train use point-pair correspondences that are generated between overhead images of the fabric in different configurations. We generate deformation-invariant correspondences by training dense object descriptors~\cite{priya-rope, dense-obj-nets} on synthetically generated images of the fabric in different configurations.
\begin{figure}[t!]
\center
\includegraphics[width=\columnwidth]{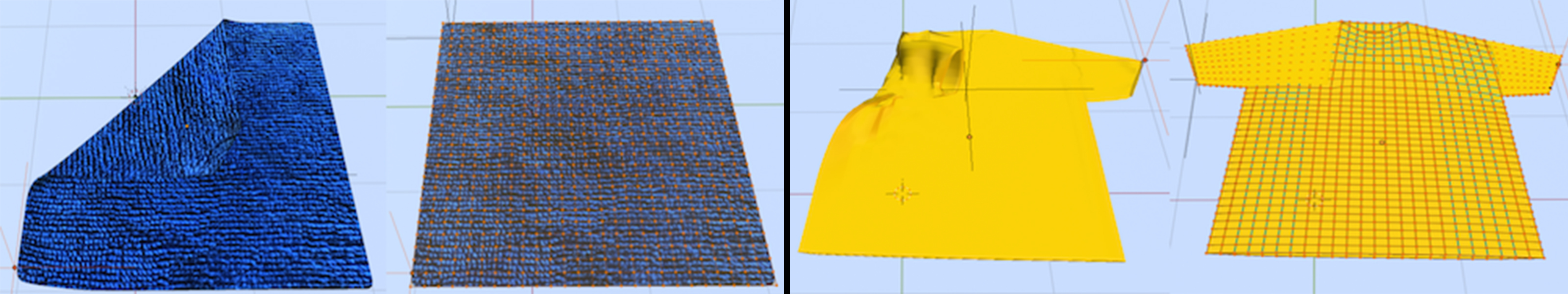}
\caption{
\textbf{Fabric Meshes: }Examples of the meshes generated in Blender for both square cloth (left) and t-shirts (right). The ground-truth vertices are highlighted in the second and fourth columns.
}
\label{fig:blender}
\end{figure}

\begin{figure}[t!]
\center
\includegraphics[width=1.0\columnwidth]{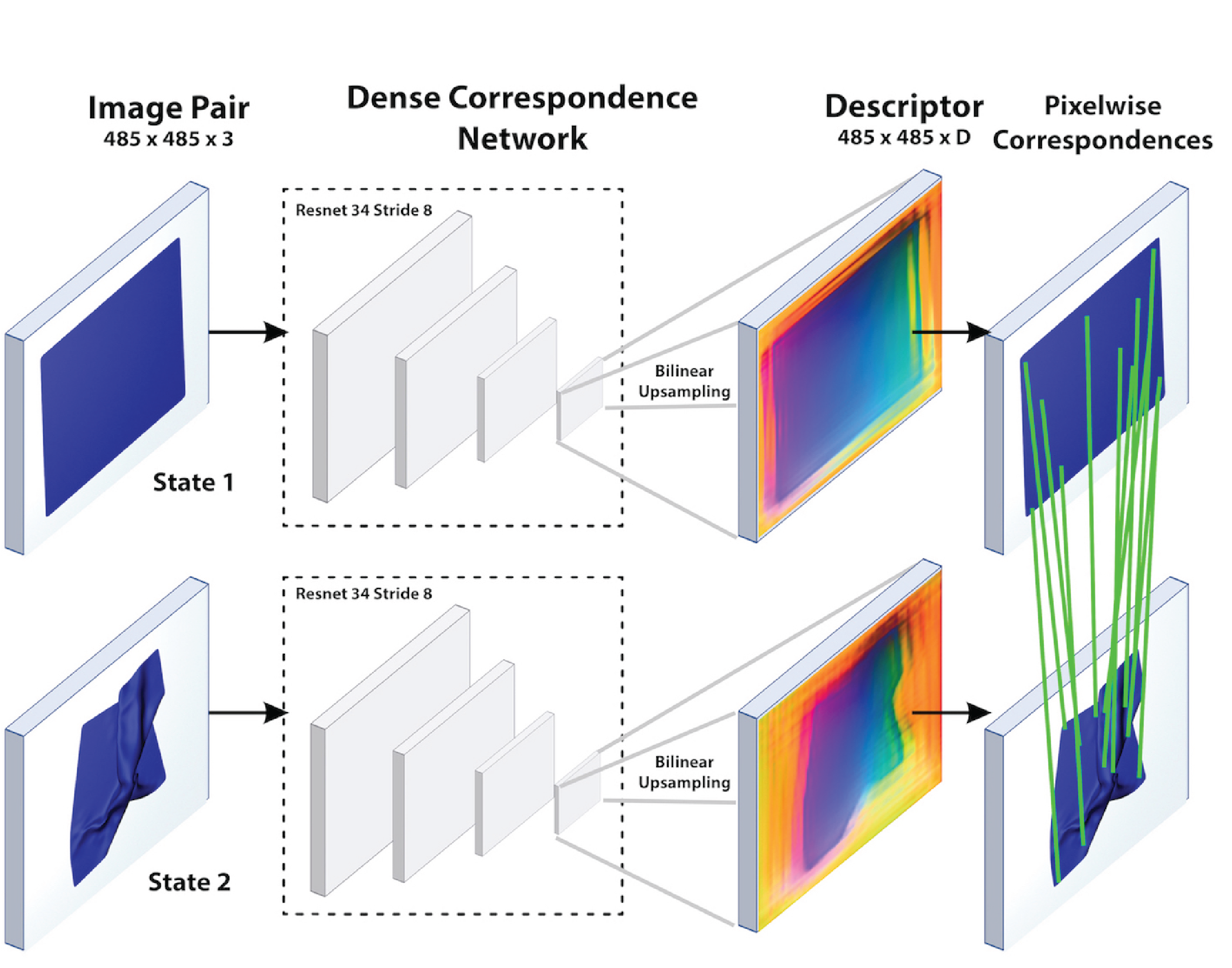}
\caption{
\textbf{Learning Visual Correspondences: }pipeline for training dense object nets for robot fabric manipulation. Left: we train a dense correspondence network on pairs of simulated fabric images to learn pixel-wise correspondences using a pixel-wise contrastive loss. Right: we use the learned descriptors for policy optimization. We can use correspondence to map a reference action to a new fabric configuration. For example, we show an image of a wrinkled fabric in ``State 2,'' and we can use descriptors to figure out the action needed to smooth the fabric from ``State 2'' to ``State 1.''
}
\label{fig:descriptors}
\end{figure}

In~\citet{dense-obj-nets}, an input image $I$ is mapped to a descriptor volume $Z = f_\theta(I) \in \mathbb{R}^{W\times H\times D}$, where each pixel $(i,j)$ has a corresponding descriptor vector $Z_{i,j}\in\mathbb{R}^D$. Descriptors are generated by a Siamese network $f_\theta$ and are guided closer together for corresponding pixels in images and pushed apart by at least some margin $M$ for non-corresponding pairs by minimizing a pixel-wise contrastive loss function during training~\cite{dense-obj-nets}. Corresponding pairs of pixels represent the same point on an object. In this work, we also train a Siamese network to output descriptors that are close for corresponding pairs of pixels and separated for non-corresponding pixel pairs for overhead images of fabric in different configurations. Since ground-truth pixel correspondences are difficult to obtain in images across deformations of a real fabric, we train the network on synthetic RGB data from Blender (see Section~\ref{sec:simulator}), where perfect information about the pixel correspondences is available through the global coordinates of the fabric mesh's vertices. Note that during training, the sampled image inputs to the Siamese network are of the same fabric type to ensure valid correspondences. That is, two different images of T-shirts can be passed into the network, but not a T-shirt and square fabric. Figure~\ref{fig:descriptors} demonstrates the pipeline for predicting descriptors for correspondence generation. The learned descriptors can then be used to approximate the correspondence function $d_{I \rightarrow  I'}$ described in Section~\ref{sec:problem_definition}:
\begin{align*}
    \left((i''_l,j''_l)\right)_{l=1}^k &= \argmin_{(i'_1,j'_1)...(i'_k, j'_k)} \sum_{l=1}^{k} \lVert f_\theta(I)_{i,j} - f_\theta(I')_{i'_l,j'_l}\rVert_2\\
    &\text{s.t. } (i'_n,j'_n) \neq (i'_m,j'_m)\; \forall m,n\in [k]\\
    d_{I\rightarrow I'}(i,j) &= \argmin_{(i', j')} \sum_{l=1}^{k} \lVert (i',j') - (i''_l, j''_l)\rVert_2
\end{align*}

The first equation computes the top $k$ pixel matches based on their distance in descriptor space and the second equation computes the geometric median of these matches in pixel space. In the physical setup, we experimentally found $k=20$ to give us the most robust predictions. %

\subsection{Dataset Generation and Domain Randomization}
\label{subsec:dataset_gen}

To enable generalization of the learned descriptors to a range of fabric manipulation tasks, we generate a diverse dataset of initial fabric configurations. The first step simulates dropping the fabric onto the planar workspace while executing similar pinning actions to those described in Section~\ref{sec:simulator} on an arbitrary subset of vertices, causing some vertices to fall due to gravity while others stay fixed. We then release the pinned vertices 30 frames later so that they collapse on top of the fabric. This allows us to create realistic deformations in the mesh. We then export RGB images which serve as inputs to the Siamese network, pixel-wise annotations which gives us correspondences, and segmentation masks which allow us to sample matches on the fabric.

Simulating soft-body animations is in general a computationally time-consuming process which makes it difficult to render large datasets in short periods of time. We take steps toward mitigating this issue by rendering 10 images per drop, allowing us to collect 10x as much data in the same time period. In simulation, we found that the test time pixel match error was unaffected when including these unsettled images of the fabric in the dataset. We additionally make use of domain randomization~\cite{cad2rl,domain_randomization} by rendering images of the scene while randomizing parameters including mesh size, lighting, camera pose, texture, color and specularity (see supplement for further details). We also restrict the rotation about the z-axis to be between $(-\pi/4, \pi/4)$ radians to reduce ambiguity during descriptor training due to the natural symmetry of fabrics such as squares. To randomize the image background, we sample an image from MSCOCO~\cite{mscoco} and ``paste'' the rendered fabric mask on top. For our experiments, we generated one (domain-randomized) dataset, including both T-shirts and square fabric, and train a single model which we used for the experiments in Section~\ref{sec:results}. For reference, generating a single dataset of 7,500 images, half T-shirts and half square cloth, with 729 annotations per image takes approximately 2 hours on a 2.6GHz 6-core Intel Core i7 MacBook Pro. 

\section{Descriptor-Parameterized Controller}
\label{sec:policy_design}

As discussed in Section~\ref{subsec:task_definition}, the robot receives a demonstration of the task consisting of actions $\left( \ba_j \right)_{j=1}^{n}$ and observations $\left( I_j \right)_{j=1}^{n}$. At execution time, the robot starts with the fabric in a different configuration, and the fabric itself may have a different texture or color. At time $j\in[n]$, the robot observes $I'_j$ then executes $\pi_j(I_j') = \left(d_{I_j \rightarrow I_j'}(  x_g, y_g )_j, \;\; d_{I_j \rightarrow I_j'}(  x_p, y_p )_j\right)$ where $d_{I_j \rightarrow I_j'}$ is defined in Section~\ref{subsec:preliminaries}. $\pi_j$ uses the geometric structure learned by the descriptor network to identify semantically relevant pixels in $I_j'$ to generate actions that manipulate these keypoints. Because the descriptor network learns about task-agnostic fabric geometry, we train one network for a variety of tasks and use it across different fabric configurations from the ones supplied in demonstrations. Additionally, we do not require separate networks for each fabric type we wish to train.

For example, one step of a task could involve grasping the top-right corner of the fabric and taking an action to place it in alignment with the bottom-left corner, thereby folding the fabric. The robot could receive an offline demonstration of this task on an initially flat fabric, but then be asked to perform the same task on a crumpled, rotated fabric. To do this, the robot must be able to identify the corresponding points in the new fabric configuration (top-right and bottom-left corners) and define a new action to align them. $\pi_j$ computes correspondences for the grasp and place points across the demonstration frame and the new observation to generate a corresponding action for the new configuration.

\subsection{Fabric Smoothing}
\label{subsec:smoothing}
In the square fabric smoothing task, the robot starts with a crumpled fabric and spreads it into a smooth configuration on a planar workspace as in~\citet{seita_ryan}. To complete this task, we use the approach from~\cite{seita_ryan} and iterate over fabric corners, pulling each one to their target locations on an underlying plane. However, while~\cite{seita_ryan} design a policy to do this using ground-truth knowledge of the fabric in simulation, we alternatively locate corners on the crumpled fabric using a learned descriptor network and a source image of a flat fabric where the corners are labeled. For the T-shirt smoothing task, we apply a similar method but instead iterate over the corners of the sleeves and the base of the T-shirt.
\subsection{Fabric Folding}
\label{subsec:method1}
The fabric folding task involves executing a sequence of folds on a fairly smooth starting configuration. For each folding task, we use a single offline demonstration containing up to 4 pick and place actions, each defined by a pick and drop pixel location and collected by a human via a GUI. The descriptor-parameterized controller is then executed in an open-loop manner.%

\begin{figure}[t!]
\center
\vspace{0.13cm}\includegraphics[width=1.0\columnwidth]{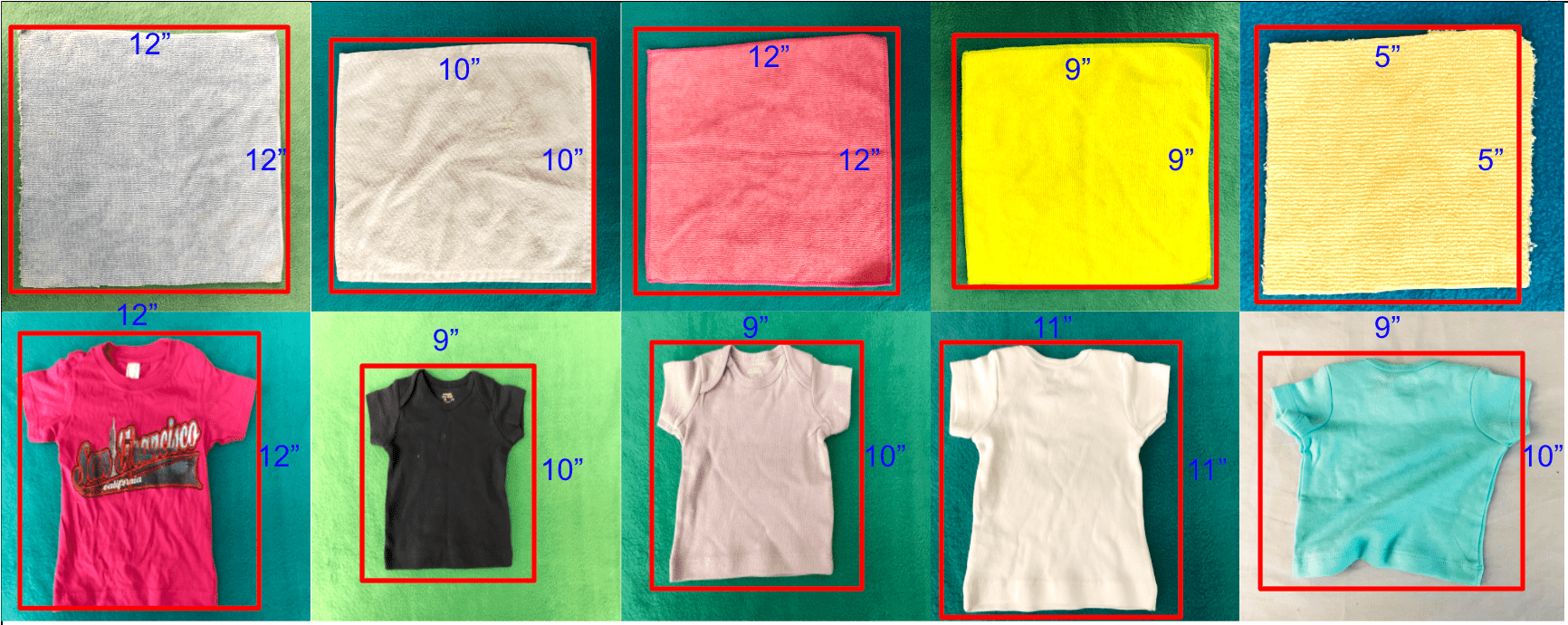}
\caption{
\textbf{Fabric Specifications: }Images and dimensions of the square fabrics and shirts we use in experiments.
}
\label{fig:fabric_specs}
\end{figure}

\section{Experiments}
\label{sec:results}
We experimentally evaluate (1) the quality of the learned descriptors and their sensitivity to training parameters and (2) the performance of the descriptor-based policies from Section~\ref{sec:policy_design} on two physical robotic systems, the da Vinci Research Kit (dVRK)~\cite{dvrk2014} and the ABB YuMi. Results suggest that the learned descriptors and the resulting policies are robust to changes in fabric configuration and color. 

\subsection{Tasks}
\label{subsec:tasks}
We consider 10 fabric manipulation tasks executed on a set of 5 T-shirts and 5 square fabrics in the real world:
\begin{enumerate}
    \item \textit{Single Fold (SF):} A single fold where one corner is pulled to its opposing corner.
    \item \textit{Double Inward Fold (DIF):} Two opposing corners are folded to the center of the fabric.
    \item \textit{Double Triangle Fold (DTF):} Two sets of opposing corners are aligned with each other.
    \item \textit{Double Straight Fold (DSF):} The square cloth is folded in half twice, first along the horizontal bisector and then along the vertical bisector. 
    \item \textit{Four Corners Inward Fold (FCIF):} All four corners are sequentially folded to the center of the cloth.
    \item \textit{T-Shirt Sleeves Fold (TSF):} The two sleeves of a t-shirt are folded to the center of the shirt.
    \item \textit{T-Shirt Sleeve to Sleeve Fold (TSTSF):} The left sleeve of a T-shirt is folded to the right sleeve of the T-shirt.
    \item \textit{Smoothing (S):} Fabric is flattened from a crumpled state.
    \item \textit{Smoothing + Double Triangle Fold (SDTF):} Fabric is smoothed then the DTF is executed.
    \item \textit{Smoothing + Sleeve to Sleeve Fold (SSTSF):} T-shirt is smoothed then TSTSF is executed.
\end{enumerate}
All fabrics are varied either in dimension or color according to Figure~\ref{fig:fabric_specs}. Additionally, we execute a subset of these tasks in simulation. A single visual demonstration consisting of up to 4 actions is provided to generate a policy which the robot then tries to emulate in the same number of actions.
\begin{figure}[t!]
\center
\includegraphics[width=\columnwidth]{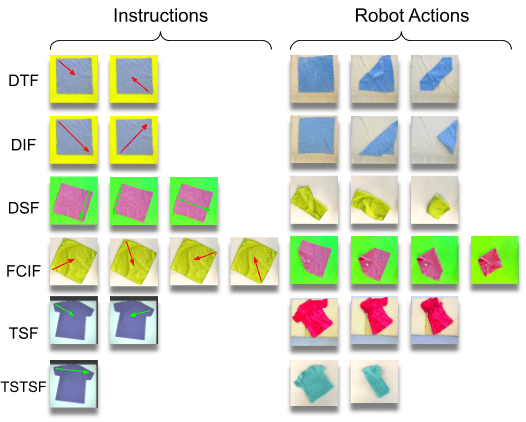}
\caption{
\textbf{Policy Rollouts: } We visualize policy execution on the YuMi for tasks 2, 3, 4, 5, 6 and 7 as described in Section~\ref{subsec:tasks}. The first four columns show the folding instructions on some initial fabric and the last four columns show the corresponding folds executed on novel starting configurations for a different fabric.
}
\label{fig:results_real}
\end{figure}

\begin{figure*}[!t]
\center
\vspace{0.13cm}\includegraphics[width=0.98\textwidth]{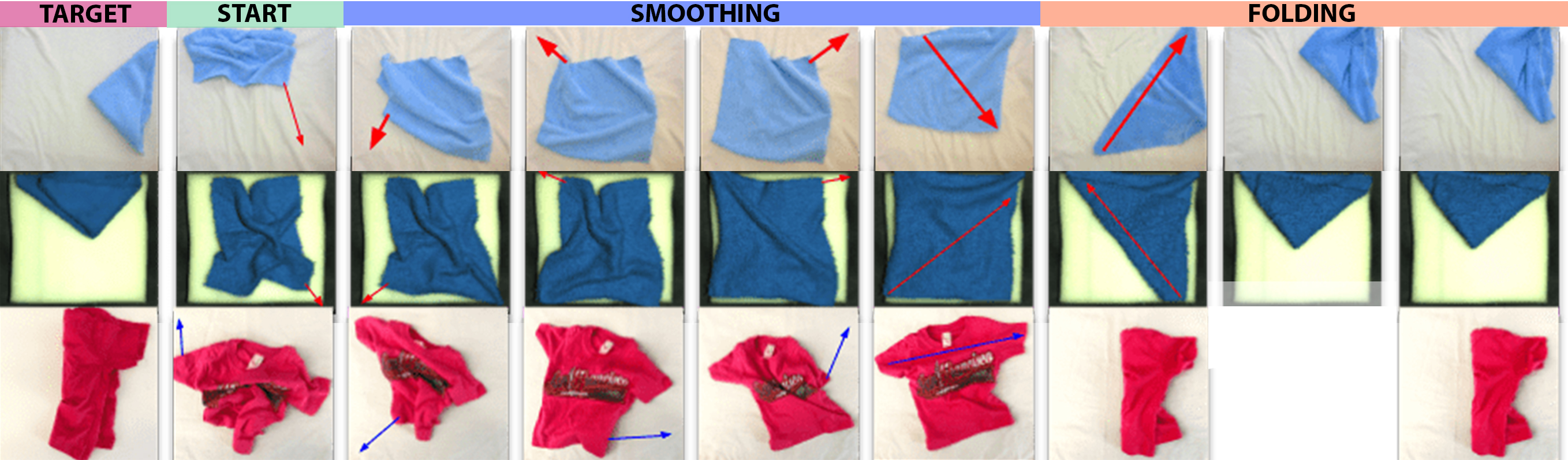}
\caption{
\textbf{Full Folding Sequence: }The first and second row is a time-lapse of a sequence of 6 actions taken by the YuMi and dVRK respectively, and with actions overlaid by red arrows, to successively smooth a wrinkled fabric and then fold it according to task 3 in Section~\ref{subsec:tasks}. The third row is a time-lapse of a sequence of 5 actions taken by the YuMi to complete task 10 in Section~\ref{subsec:tasks}. Here, robot actions are overlaid with blue arrows.
}
\label{fig:results_smooth+fold}
\end{figure*}

\subsection{Experimental Setup}
\label{physical_exps_details}
We execute fabric folding and smoothing experiments on the dVRK~\cite{dvrk2014} and ABB YuMi robot. The dVRK is equipped with the Zivid OnePlus RGBD sensor that outputs $1900\times1200$ pixel images at 13 FPS at depth resolution $0.5$ mm. The workspace of the dVRK is only $5"\times5"$, so we use only square fabric of the same dimension while varying the color according to Figure~\ref{fig:fabric_specs}. Manipulating small pieces of fabric into folds is challenging due to the elasticity of the fabric, so we add weight to the fabric by dampening it with water. Additionally, we place a layer of 1 inch foam rubber below the fabric to avoid damaging the gripper. The YuMi has a $36"\times24"$ workspace, and since only one arm is utilized resulting in a more limited range of motion, we only manipulate at most $12"\times12"$ pieces of fabric which we do not dampen. In this setup we use a 1080p Logitech webcam to collect overhead color images. For the YuMi, we use both T-shirts and square fabric of varying dimension and color but go no lower than $9"\times9"$ fabrics due to its larger gripper. Finally, for both robots, we use a standard pixel to world calibration procedure to get the transformation from pixel coordinates to planar workspace coordinates.

For both robots, we follow the same experimental protocol. We manually place the fabric in configurations similar to those shown in Figure~\ref{fig:blender} and deform them by pulling at multiple locations on the fabric. To obtain image input for the descriptor networks, we crop and resize the overhead image to be $485\times485$ such that the fabric is completely contained within the image. Although lighting conditions, camera pose and workspace dimensions are significantly different between the two robotic systems, no manual changes are made to the physical setup. We find that the learned descriptors are sufficiently robust to handle this environmental variability.

We evaluate the smoothing task by computing the coverage of the cropped workspace before and after execution. For the folding tasks, as in~\citet{fabric_folding_real}, we consider an outcome a success if the final state is visually consistent with the goal image. Conventional quantitative metrics such as intersection of union between the final state and a target image provide limited diagnostic information when starting configurations are significantly different as in the presented experiments.

\subsection{Results}
\label{subsec:physical_exps}
We evaluate the smoothing and folding policies on both the YuMi and dVRK on square fabrics and T-shirts. Table~\ref{tab:phys_results} shows the success rates of our method on all proposed tasks in addition to a breakdown of the failure cases detailed in Table~\ref{tab:failures}. We observe that the descriptor-parameterized controller is able to successfully complete almost all folding tasks at least $75\%$ of the time, and the smoothing policies are able to increase coverage of the cloth to over $83\%$ (Table~\ref{tab:phys_results_smoothing}). The execution of the smoothing policy followed by the double triangle folding policy results in successful task completion $6/10$ and $8/10$ times on the YuMi and dVRK respectively. We find that the most frequent failure mode is an unsuccessful grasp of the fabric which is compounded for tasks that require more actions. Though this is independent of the quality of the learned descriptors, it highlights the need for more robust methods to grasp highly deformable objects.  

\begin{table}[!htbp]
\centering
\resizebox{\columnwidth}{!}{
 \begin{tabular}{||l || c || r || r || r ||} 
 \hline
 Task & Robot & Avg. Start Coverage & Avg. End Coverage  \\ 
 \hline\hline
 S & YuMi & $71.4 \pm 6.2$ & $83.2 \pm 8.1$\\
 \hline
 S & dVRK & $68.4 \pm 4.4$ & $86.4 \pm 5.2$\\
\hline
\end{tabular}}
\caption{\textbf{Physical Fabric Smoothing Experiments:} We test the smoothing policies designed in Section~\ref{sec:policy_design} on the YuMi and the dVRK robots. Both robots are able to increase coverage during the smoothing task by $11-22$ percent on average.}
\label{tab:phys_results_smoothing}
\end{table}

\begin{table}[!htbp]
\centering
\resizebox{\columnwidth}{!}{
 \begin{tabular}{||l || c || c || r || r || r || r ||} 
 \hline
 Task & Robot & \# Actions & Success & Error A & Error B & Error C\\ 
 \hline\hline
 SF & YuMi & 1 & 18/20 & 2 & 0 & 0\\
 \hline
 SF & dVRK & 1 & 20/20 & 0 & 0 & 0\\
 \hline
 DIF & YuMi & 2 & 16/20 & 3 & 0 & 1\\
 \hline
 DIF & dVRK & 2 & 20/20 & 0 & 0 & 0\\
 \hline
 DTF & YuMi & 2 & 14/20 & 3 & 2 & 1\\
 \hline
 DTF & dVRK & 2 & 18/20 & 0 & 2 & 0\\
 \hline
 TSF & YuMi & 2 & 15/20 & 3 & 0 & 2\\
 \hline
 SDTF & YuMi & 6 & 6/10 & 2 & 1 & 1\\
 \hline
 SDTF & dVRK & 6 & 8/10 & 0 & 2 & 0\\
 \hline
 DSF & YuMi & 3 & 15/20 & 1 & 1 & 3\\
 \hline
 DSF & dVRK & 3 & 17/20 & 1 & 0 & 2\\
 \hline
 FCIF & YuMi & 4 & 13/20 & 5 & 1 & 1\\
 \hline
 FCIF & dVRK & 4 & 18/20 & 0 & 1 & 1\\
 \hline
 TSTSF & YuMi & 1 & 17/20 & 2 & 0 & 1\\
 \hline
 SSTSF & YuMi & 5 & 6/10 & 2 & 0 & 2\\
 \hline
\end{tabular}}
\caption{\textbf{Physical Fabric Folding Experiments:} We test the folding policies from Section~\ref{sec:policy_design} on the YuMi and the dVRK. We observe both robots are able to perform almost all folding tasks at least $75$ percent of the time. The YuMi is able to perform the smoothing then folding task $6/10$ times and the dVRK is able to do so $8/10$ times.}
\label{tab:phys_results}
\end{table}

\begin{table}[!htbp]
\centering
\resizebox{\columnwidth}{!}{
 \begin{tabular}{||c || p{0.3\textwidth} ||} 
 \hline
 Error & Description\\ 
 \hline\hline
 A & Gripper picks up more than one layer of fabric or fabric slips out of gripper due to inaccurate depth of grasp\\
 \hline
 B & Pick or drop correspondence error greater than 30 pixels (10\% of cloth width) or pick correspondence not on fabric mask\\
 \hline
 C & Unintended physics: resulting fold does not hold due to variable stiffness of the fabric, friction of the fabric, or friction of the underlying plane\\
 \hline
\end{tabular}}
\caption{\textbf{Failure Mode Categorization}}
\label{tab:failures}
\end{table}
\section{Discussion and Future Work}
\vspace{-0.05in}
\label{sec:discussion}
We present an approach for multi-task fabric manipulation by using visual correspondences learned entirely in simulation. Experiments suggest that the learned correspondences are robust to different fabric colors, shapes, textures, and sizes and make it possible to efficiently learn 10 different fabric smoothing and folding tasks on two different physical robotic systems with no training in the real world. In future work, we plan to explore hierarchical fabric manipulation policies, where visual correspondences can be used to define coarse action plans while a closed loop controller can be learned to realize these plans. We will also explore more complex fabric manipulation tasks, such as wrapping rigid objects, in which reasoning about fabric dynamics is critical or tasks involving manipulating multiple fabrics simultaneously.

\printbibliography
\clearpage

 \normalsize
 \section{Appendix}
 \label{sec:appendix}
 The appendix is organized as follows:
 \begin{itemize}
     \item Appendix A contains additional details on the fabric simulator
     \item Appendix B contains additional details on the experiments conducted in simulation
     \item Appendix C contains results for simulation experiments
     \item Appendix D shows visualizations of the learned descriptor mappings.
     \item Appendix E contains images from additional physical trials executed on the dVRK and YuMi.
     \item Appendix F conducts a detailed study on the effect of various hyperparameters on the quality of the learned visual correspondences.
     
 \end{itemize}
 \subsection{Fabric Simulator Details}
\begin{figure}[b]
 \center
 \includegraphics[width=0.48\textwidth]{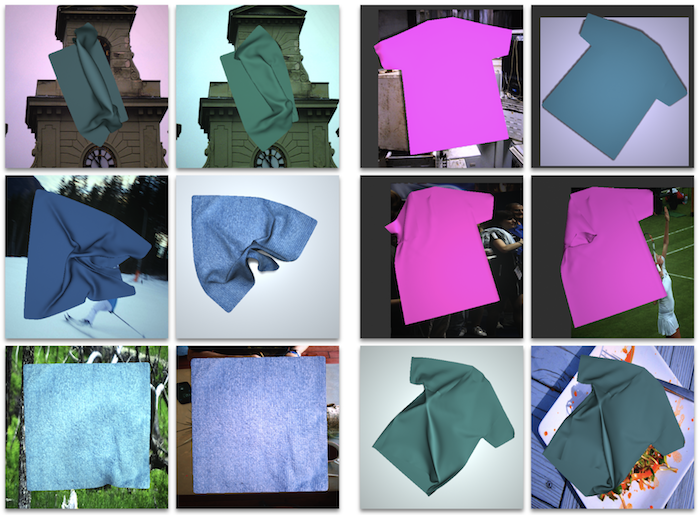}
 \caption{
 \small
 Examples of domain-randomized images of the starting fabric states encountered in the dataset generation phase described in Section~\ref{subsec:dataset_gen}. The first two columns show examples of images with a square fabric, and the last two columns show similar examples but with a t-shirt.
 }
 \label{fig:dom-rand}
 \end{figure}

 We use Blender 2.8 to both create dynamic cloth simulations and to render images of the fabric in different configurations. As can be seen in Figure~\ref{fig:blender}, we are able to retrieve the world coordinates of each vertex via Blender's API which we then use to find ground truth pixel correspondences through an inverse camera to world transformation. This allows us to create dense pixel-vertex annotations along the surface of the fabric which we feed to into the descriptor training procedure. Figure~\ref{fig:descriptor_vis_res} is a visualization of the learned descriptors and Figure~\ref{fig:dom-rand} contains examples of the domain randomized training data we generate through Blender.
 
 \subsubsection{Fabric Model}
\begin{figure}[h]
 \center
 \includegraphics[width=0.48\textwidth]{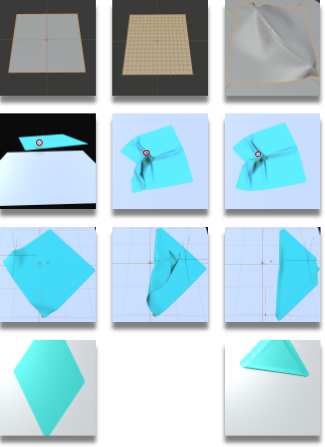}
 \caption{
 \small
 The top row illustrates the the process of creating cloth in Blender from a default square mesh. The second row is an example of a starting configuration generated by dropping the cloth from a fixed height and pinning a single arbitrary vertex. The pinned vertex is labeled by the red circle. The third row illustrates frames from a folding action in the simulator and the last row shows the corresponding rendered images of the settled cloth before and after the action.
 }
 \label{fig:start_config}
 \end{figure}

 To generate the square cloth in Blender, we first import a default square mesh and subdivide it three times to create a grid of $27\times27$ grid of vertices. We found that this number of square vertices resulted in a visually realistic animation in comparison to our real fabrics. We additionally add 0.02 meter thickness to the cloth to increase its weight which creates more realistic collision physics. In order to apply Blender's in-built cloth physics to the mesh, we simply make use of the cloth physics modifier through which we are able modify the parameters shown in Table~\ref{tab:simulator_params}. Internally, Blender simulates fabric physics for polygonal meshes with gravitational forces, damping, stiffness, and by interconnecting the mesh vertices with four types of virtual springs: tension springs, compression springs, shear springs, and angular bending springs. Each vertex also exerts repulsive forces within a self-contained virtual sphere on vertices both within fabric and in surrounding objects, to simulate self-collisions and collisions with other objects. We visually tune the simulator by replaying a fabric folding action while varying parameters, most notably the friction coefficients and spring elasticity constants. From observing videos of the folding actions, we settle on the parameter values specified in Table \ref{tab:simulator_params}. A visualization of these steps can be seen in the top row of Figure~\ref{fig:start_config}. To generate the t-shirt mesh, we similarly import a default square mesh and subdivide it three times, but also delete all vertices that do not lie in a predefined t-shirt cutout of the square mesh which results in the bottom right image of Figure~\ref{fig:blender}. 
 \subsubsection{Manipulation with Hook Objects}
 We utilize hook objects to take actions in the Blender simulator. A hook object attaches to a mesh vertex and exerts a proportional sphere of influence over the selected vertex and those in its vicinity, pulling the fabric in the direction of movement. We simulate a grasp, drag, and drop of the fabric by assigning a hook object to a fabric vertex, moving this hook over a series of frames to the deprojected pixel drop location, and removing the hook object assignment to release the cloth.

 \begin{table}[!t]
    
    \caption{Blender Cloth Simulation Parameters}
    \centering
    \resizebox{\columnwidth}{!}{
    \begin{tabular}{||l|| c|| r||} 

 \hline
 Parameter & Explanation & Value  \\ 
 \hline\hline
  Quality Steps & quality of cloth stability and collision response  & 5.0 \\
 \hline
 Speed Multiplier & how fast simulation progresses  & 1.0 \\
 \hline
 Cloth Mass (kg) & --  & 0.3 \\
 \hline
 Air Viscosity & air damping & 1.0 \\
 \hline
 Tension Springs & tension damping/stretching  & 5.0 \\
 \hline
 Compression Springs & compression damping/stretching  & 5.0 \\
 \hline
 Shear Springs & damping of shear behavior  & 5.0 \\
 \hline
 Bending Springs & damping of bending behavior  & 0.5 \\
 \hline
 Friction & friction with self-contact  & 5 \\
 \hline
 Self-Collision Distance (m) & per-vertex spherical radius for repulsive forces  & 0.015 \\
 \hline
\end{tabular}
}
\label{tab:simulator_params}
\end{table}
 
 \begin{figure}[htb!]
 \center
 \includegraphics[width=0.48\textwidth]{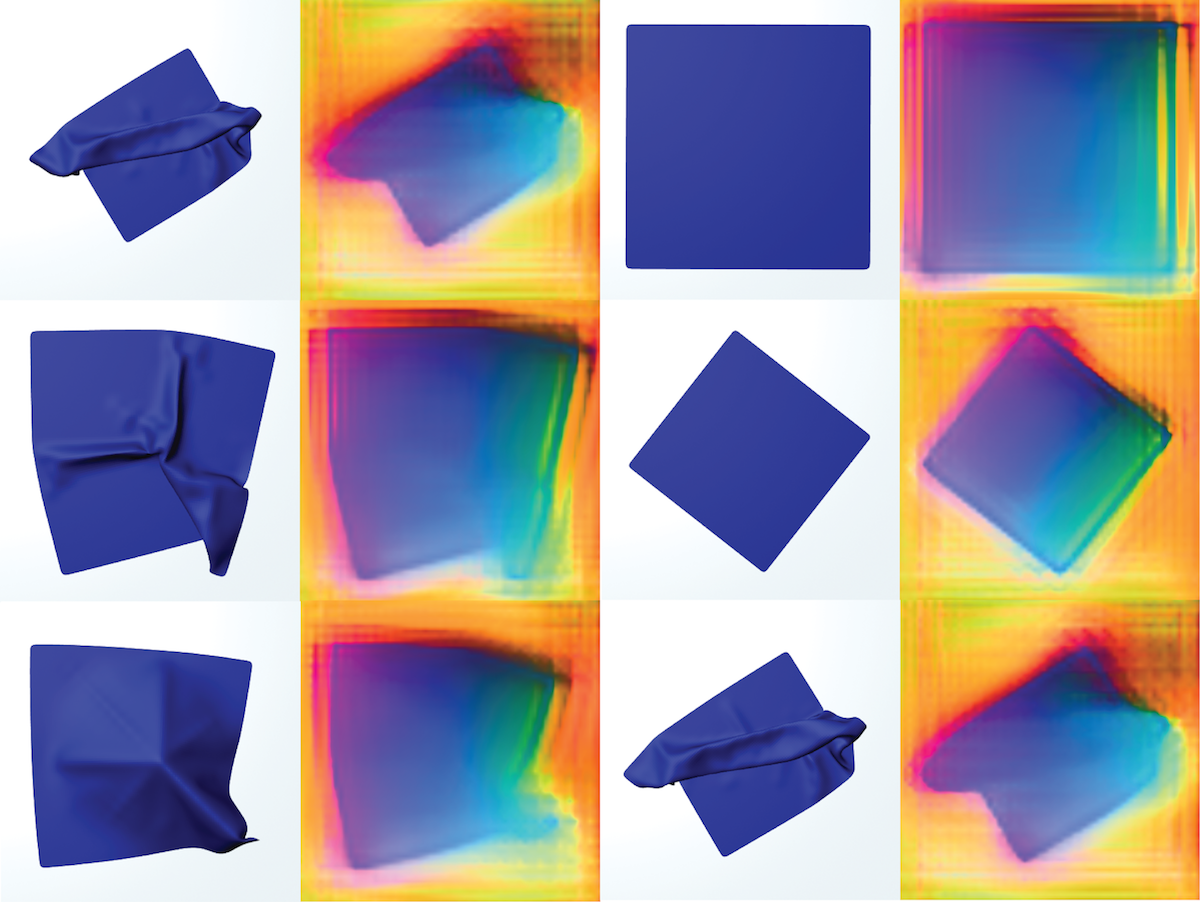}
 \caption{
 \small
 Visualization of the 3-dimensional descriptors learned via the training procedure described in Section~\ref{sec:descriptors} by mapping each pixel's descriptor vector to an RGB vector. Thus, similar colors across the images of columns two and four represent corresponding points on the square cloth.
 }
 \label{fig:descriptor_vis_res}
 \end{figure}
 
\begin{figure*}[h]
 \center
 \includegraphics[width=\textwidth]{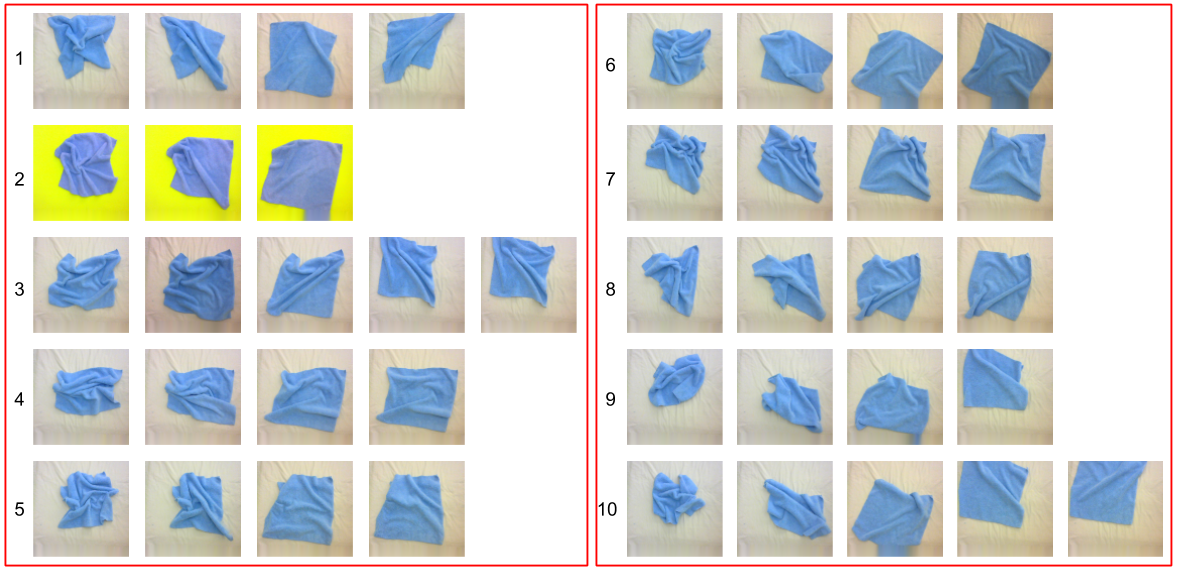}
 \caption{
 \small
 Additional rollouts of the smoothing task from randomly chosen starting configurations. The learned descriptors are used to locate the corners of the fabric and successively pull them to a reference location in an image of the flat cloth.
 }
 \label{fig:smoothing-exps}
 \end{figure*}
 
 \begin{figure*}[h]
 \center
 \includegraphics[width=\textwidth,height=13.0cm]{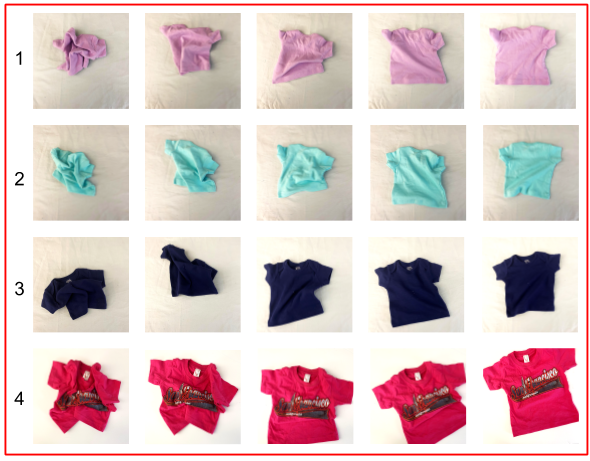}
 \caption{
 \small
 Additional rollouts of the smoothing task from randomly chosen starting configurations. The learned descriptors are used to locate the corners of the fabric and successively pull them to a reference location in an image of the flat cloth.
 }
 \label{fig:tshirt-smoothing-exps}
 \end{figure*}
 
\begin{figure*}[h]
 \center
 \includegraphics[width=\textwidth]{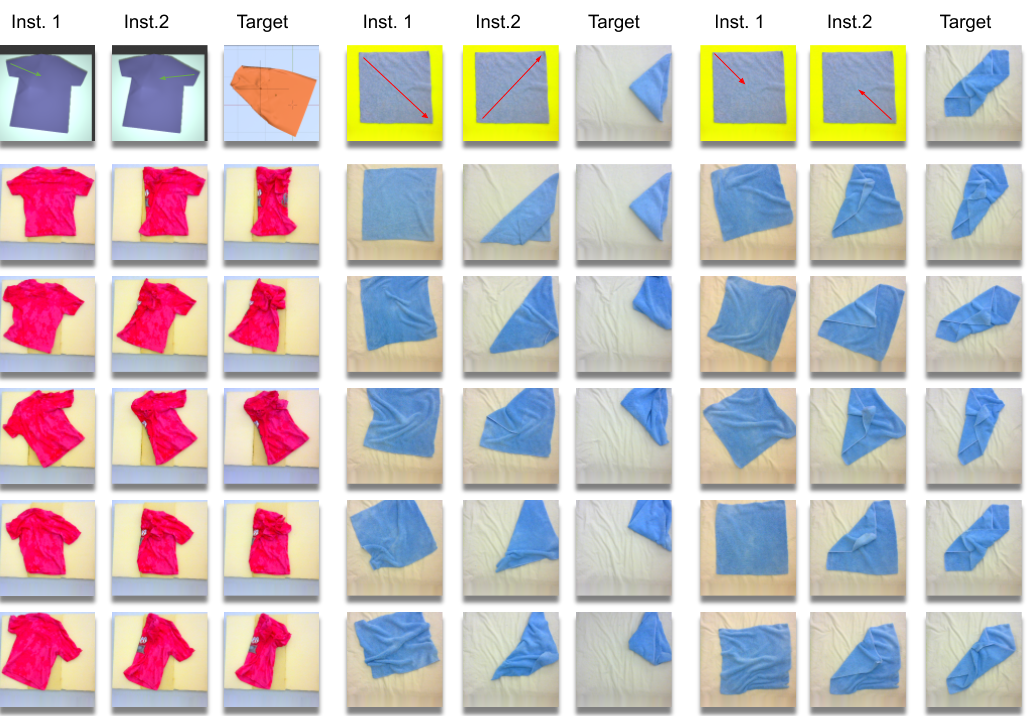}
 \caption{
 \small
 Additional rollouts of the folding tasks described in Section~\ref{subsec:tasks} from arbitrary starting configurations. The left column, center column and right column contain results for tasks 4, 3 and 2 respectively.
 }
 \label{fig:folding-exps}
 \end{figure*}
  
 \subsubsection{Starting Configurations and Actions}
 To generate varied starting configurations, we simulate dropping the fabric from 0.2 meters above the workspace while pinning it an arbitrary subset of the 729 vertices. After 30 frames in the animation, the pinned vertices are released and are allowed to settle for another 30 frames. This creates natural deformation in the cloth and introduces a wide range of starting configurations to the training dataset. A sequence of these steps is shown in the second row of Figure~\ref{fig:start_config}. When running simulated experiments, taking pick and place actions requires manipulating the cloth via hook objects as defined in Section~\ref{sec:simulator}. A sequence of frames throughout the course of an action using a hook object as well as the corresponding rendered frames are shown in the last two rows of Figure~\ref{fig:start_config}.   
 
\subsection{Simulation Experiment Details}
\label{simulation_exps_detail}
In simulation, we conduct 50 trials of the first 4 folding tasks described in \ref{subsec:tasks} on a domain randomized test set generated as described in \ref{subsec:dataset_gen}. We consider an outcome a success if the final state is visually consistent with the target image. We additionally declared a failure when the planned pick and drop pixels were more than 50 pixels away from their correct ground truth locations which we had access to in Blender. Note that this is neither a sufficient nor necessary condition for a successful fold, but nevertheless serves as a decent heuristic. While we considered more quantitative metrics such as structural similarity between the target image and the final state and summed distance between corresponding vertices on the mesh, these metrics are insufficient when the test time starting configuration is significantly different from the demonstration configuration.

\subsection{Simulation Experiment Results}
\label{subsec:simulated_exps}
We evaluate the folding policies designed in Section~\ref{sec:policy_design} in the simulated fabric environment. The policies successfully complete the tasks $84$ to $96$ percent of the time (Table~\ref{tab:results_sim}).

\begin{table}[!htbp]
\centering
\resizebox{0.6 \columnwidth}{!}{
 \begin{tabular}{||l || c || r  ||} 
 \hline
 Task & Success Rate  \\ 
 \hline\hline
 Single Fold & 46/50\\
 \hline
 Double Inward Fold & 48/50\\
 \hline
 Double Triangle Fold & 42/50\\
 \hline
 T-Shirt Sleeves Fold & 44/50\\
 \hline
\end{tabular}}
\caption{\textbf{Simulated Fabric Folding Experiments:} We observe that the system is able to successfully complete the tasks $84$ to $96$ percent of the time in simulation. Success is determined by visual inspection of the cloth after the sequence of actions is executed. Example simulation rollouts are shown in Figure~\ref{fig:rollouts_sim}.}
\label{tab:results_sim}
\end{table}
 
 \subsection{Descriptor Mapping Visualizations}
 We present the descriptor volumes produced by a model trained to output 3-dimensional descriptors. We coarsely visualize the volumes by presenting them as RGB images (Figure~\ref{fig:descriptor_vis_res}), and observe that corresponding pixels of the cloth map to similar colors in the descriptor volumes across configurations.

\subsection{Physical Trial Trajectories}
In this section, we present additional trials of the physical experiments conducted using the descriptor-based policies for smoothing (Figure~\ref{fig:smoothing-exps}, Figure~\ref{fig:tshirt-smoothing-exps}) and folding (Figure~\ref{fig:folding-exps}).

\subsection{Descriptor Quality Ablations}
\label{subsec:descriptor_exps}

\begin{figure}[t!]
\center
\includegraphics[width=1.0\columnwidth]{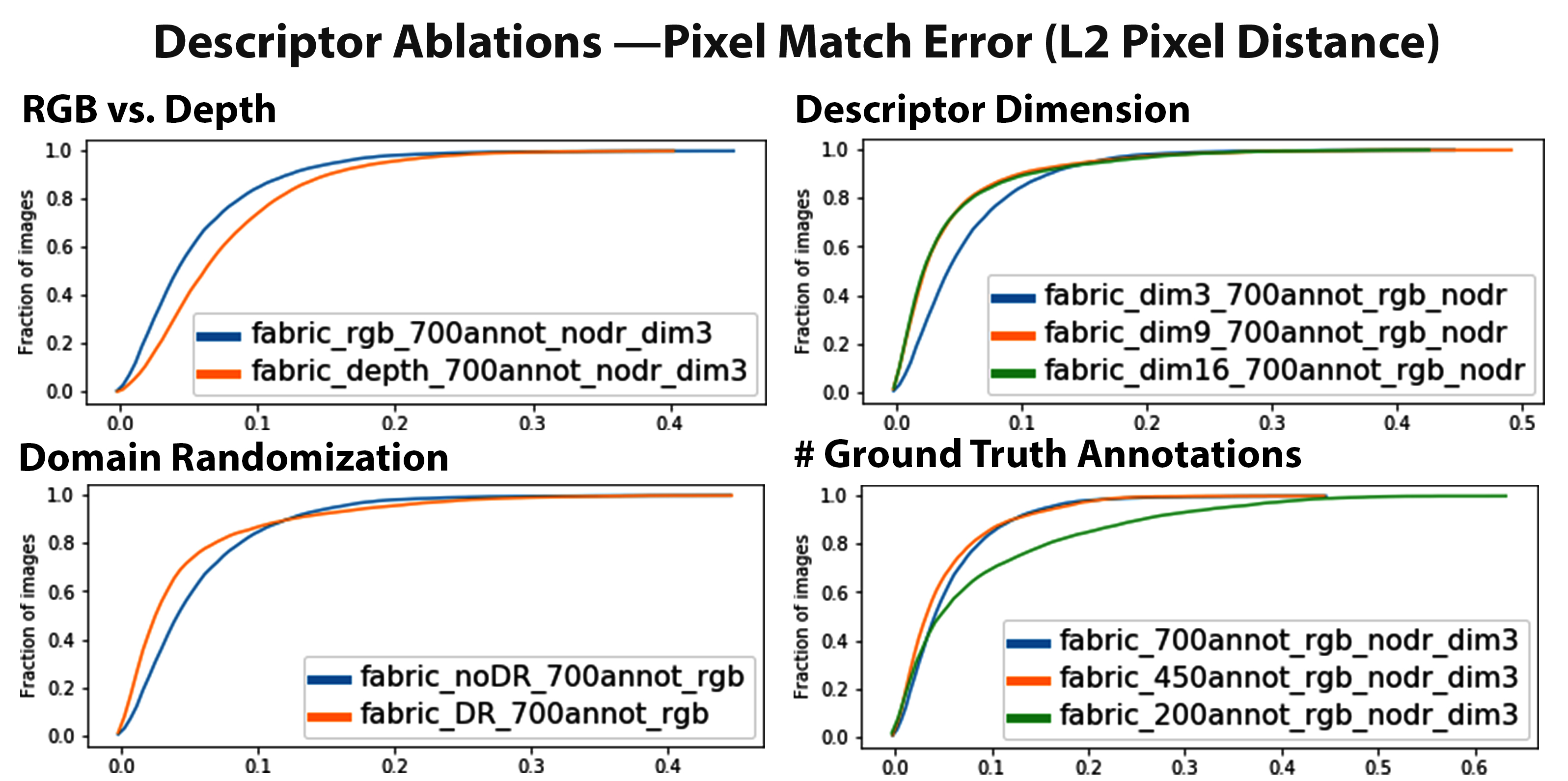}
\caption{
\textbf{Ablation studies: } We study the sensitivity of the learned dense object descriptors as described in Sections~\ref{sec:descriptors} and~\ref{subsec:descriptor_exps} to training parameters. Starting from top left, and proceeding clockwise, we test the effect of testing on RGB vs depth images, on the descriptor dimension (either 3, 9, or 16), on the number of ground truth annotations, and whether domain randomization is used. All results are evaluated using pixel match error on a held-out set of image pairs.
}
\label{fig:descriptors_results}
\end{figure}

\begin{figure}[t!]
\center
\includegraphics[width=1.0\columnwidth]{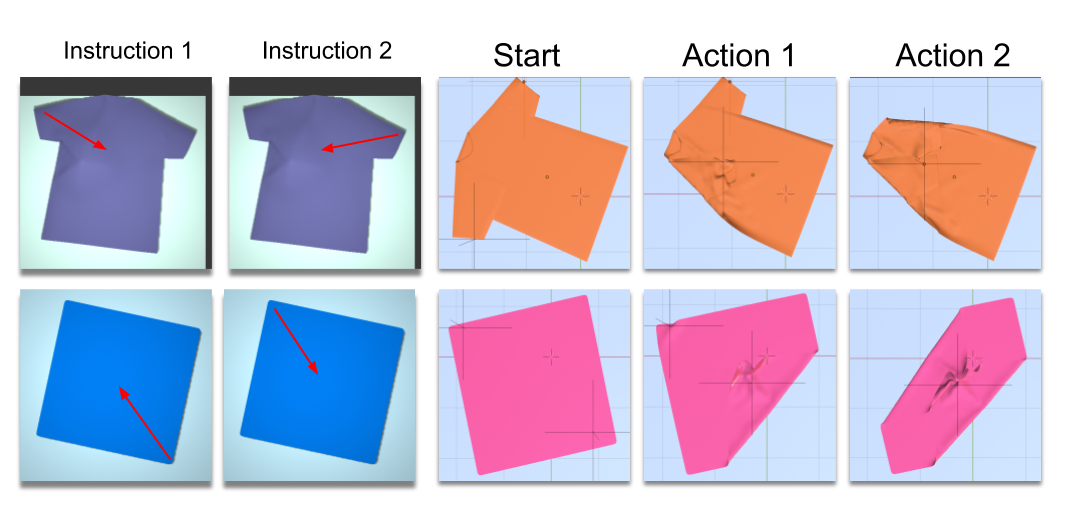}
\caption{
\textbf{Simulation Policy Visualization: }Visualization of the policy executed in simulation (with Blender) using learned descriptors for folding tasks 2 and 4 described in \ref{subsec:tasks}. The first two columns show the corresponding folding instructions from a web interface (pick-and-place actions shown with red arrows) for tasks 2 and 4. The third column shows images of the previously unseen initial configurations of fabrics before the actions, while the last two columns show the result of executing descriptor-parameterized actions. Results suggest that the learned descriptors can be used to successfully perform a variety of folding tasks from varying initial configurations.}
\label{fig:rollouts_sim}
\end{figure}

To investigate the quality of the learned descriptors with training process described in Section~\ref{sec:descriptors}, we perform four sets of ablation studies. We evaluate the quality of learned descriptors in a manner similar to~\citet{priya-rope} by evaluating the $\ell_2$ pixel distance of the pixel match error on a set of 100 pairs of held-out validation set images, where for each we sample 100 pixel pairs. We study the effect of training descriptors on (1) RGB or depth images, (2) using descriptor dimension 3, 9, or 16, (3) using 200, 450, or 700 ground-truth annotated images, and (4) whether domain randomization is used or not. Results suggest that the learned descriptors are best with RGB data, with descriptor dimension between 3 and 16 and with domain randomization, though the performance is generally insensitive to the parameter choices, suggesting a robust training procedure. Based on these results, we use RGB images with domain randomization, and with descriptor dimension 3 for all simulated experiments for both the t-shirt and square fabric. We use RGB, domain-randomized, 9-dimensional descriptors for real fabric experiments. See Figure~\ref{fig:descriptors_results} for plots.

\end{document}